# DCT-CompCNN: A Novel Image Classification Network Using JPEG Compressed DCT Coefficients

Bulla Rajesh[1], Mohammed Javed[1*], Ratnesh[2], Shubham Srivastava[2]

[1]Department of IT, Indian Institute of Information Technology, Allahabad, U.P -211012, India
[2]Department of CSE, Madan Mohan Malaviya University of Technology, Gorakhpur, U.P -273010, India
{Email: *javed@iiita.ac.in}

**Abstract:** The popularity of Convolutional Neural Network (CNN) in the field of Image Processing and Computer Vision has motivated researchers and industrialist experts across the globe to solve different challenges with high accuracy. The simplest way to train a CNN classifier is to directly feed the original RGB pixels images into the network. However, if we intend to classify images directly with its compressed data, the same approach may not work better, like in case of JPEG compressed images. This research paper investigates the issues of modifying the input representation of the JPEG compressed data, and then feeding into the CNN. The architecture is termed as DCT-CompCNN. This novel approach has shown that CNNs can also be trained with JPEG compressed DCT coefficients, and subsequently can produce a better performance in comparison with the conventional CNN approach. The efficiency of the modified input representation is tested with the existing ResNet-50 architecture and the proposed DCT-CompCNN architecture on a public image classification datasets like Dog Vs Cat and CIFAR-10 datasets, reporting a better performance.

## I. Introduction

The astonishing growth in Convolutional Neural Networks (CNN), in the past few years, in terms of performance and accuracy on variety of tasks has caught attention of many researchers [1-3][12-13][15]. Because of the popularity of CNN in image based applications, the CNN filters are commonly used in extracting relevant features and shapes from an image for the classification of objects. When CNN's are trained using image data as input, data is more often provided as an array of red-green-blue (RGB) pixels. However, when the images are compressed, the RGB channels will be available only when they are fully decompressed, which means at an additional computing cost. Since, the decompressed raw data used for processing/analytics can't be kept in the uncompressed form, the decompression and recompression operations have to be applied as many times the data needs to be processed. It would be very interesting, if these additional operations (decompression and recompression) can be avoided and a provision is made to feed the compressed data directly into the CNN. This stimulating research issue of feeding compressed data directly into the CNN for accomplishing classification is being investigated in this research paper taking the case study of JPEG compressed DCT coefficients.

In the literature, several research papers have addressed the issue of image classification and recognition using DCT coefficients with different approaches [20]. Deep learning has also played an important role in image classification, as a result of which several works have been reported for image classification using deep learning [3][4][5][19]. Generally, image classification is accomplished in two ways- supervised and unsupervised algorithms. Unsupervised algorithm for underwater fish recognition for image classification is discussed in [1], while supervised algorithm for image classification is discussed in [2]. Some other works also focused on DCT coefficients for classification. In [6], DCT coefficients were trained for classification of different texture using ANN and EFuNN. Different variants of DCT have been reported in [8]. The effectiveness of these variants were evaluated, and their discriminative features were extracted, which were used for handwritten digit recognition. Histogram of quantized 4 x 4 DCT coefficients were used in [10] for face recognition. In another approach DCT Coefficients were compressed by truncation in [11], and were trained to speed up image classification. In [12], an approach to train the convolutional network on frequency representation of data is demonstrated.

The idea of accelerating neural network training and inference speed is explored in [13], where instead of passing the RGB pixels as input to the network they modified the *libjpeg* library for JPEG partial decoding, and fed it as input to the network, and as a result they achieved faster training speed. The faster results achieved in this work, motivated us to follow the approach and apply it directly on JPEG compressed images. In this paper, we worked on two different variants of JPEG compressed coefficients- quantized and un-quantized. The compressed images were partially decoded and the DCT coefficients obtained were fed as input to the neural network separately after some transformation. The experimental results on the datasets (CIFAR-10 and Dogs vs Cats), demonstrates the performance of the proposed model. Rest of the paper is organized as follows- Section II gives proposed model, section III describes experimental results, and section IV concludes the research paper.

## II. Proposed Model

A typical model for compression and decompression of JPEG image data is shown in Figure 1. In this paper, we did an investigative study on the idea of directly feeding the JPEG compressed DCT coefficient into the CNN for image classification. Normally the JPEG images represented in the format of an array of RGB pixels are fed as an input to the neural network. But here, we propose two different approaches for training CNN with JPEG compressed data. With JPEG compressed images, our first approach would apply partial decoding and de-quantization on it as shown in Figure 2. After applying some transformation on the DCT coefficients obtained after de-quantization, the output of which is fed into the neural network. While in the second approach, since the compressed JPEG image was not quantized during compression, we have un-quantized JPEG compressed image, so we apply partial decoding on it as shown in Figure 3, and after some transformation, the obtained data is fed as input to the CNN network. We further trained and tested the model on a different dataset with these approaches, and which turns out to work rationally very well.

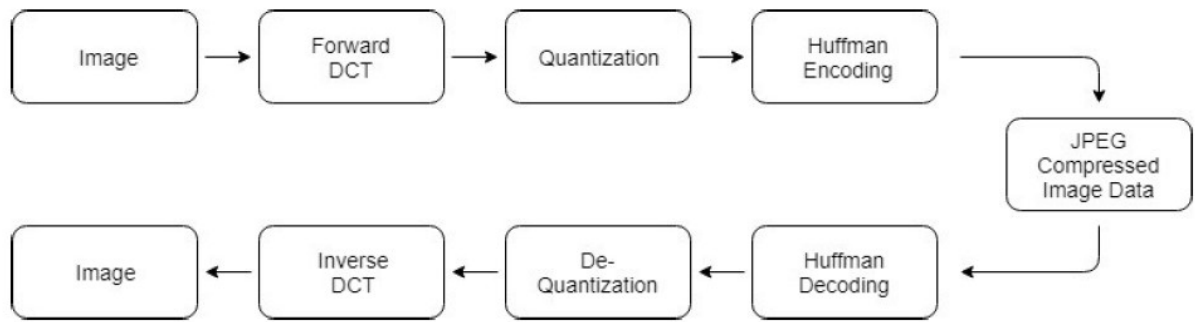

Fig. 1. Flow diagram for JPEG Compression and Decompression

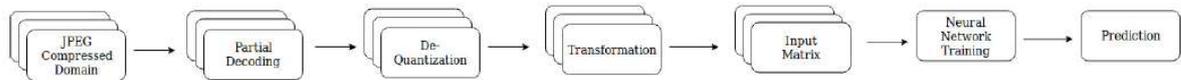

Fig. 2. First approach employed in this paper i.e. the quantized JPEG compressed image is partially decoded and de-quantized and after some transformation, is directly used as input to CNN.

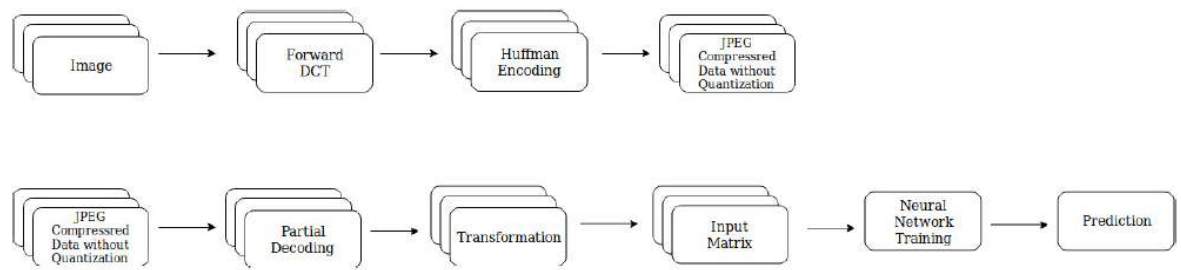

Fig. 3. Second approach employed in this paper i.e. the un-quantized JPEG compressed image is partially decoded and after some transformation, is directly used as input to CNN.

Before, discussing our proposed model lets first consider some careful design aspects. The encoded JPEG image has chroma channels dimension mismatching from that of luma channel i.e. the luma channel have the larger dimension as that of chroma channels, so it becomes necessary that their dimensions should be matching before the three channels are concatenated and fed to the network. So for matching the dimension, either we perform down-sampling on the luma channel, or up-sampling on the chroma channels. So we have analysed both the methods i.e. by down sampling the luma channel by factor of 2 and up-sampling the chroma channel by factor of 2. After performing this transformation, the three channels are concatenated and fed as input to the CNN. CNN architecture consists of different layers to process the image, and these layers are categorized into two parts i.e. feature learning and classification. The first layer is a feature extractor which extracts feature from the image and produces feature maps as output, and identically the other hidden layer extract feature from its lower level feature map. Finally, the output of these feature extractor is fed to the fully connected layer which performs classification. The proposed DCT-CompCNN is shown in Figure 4.

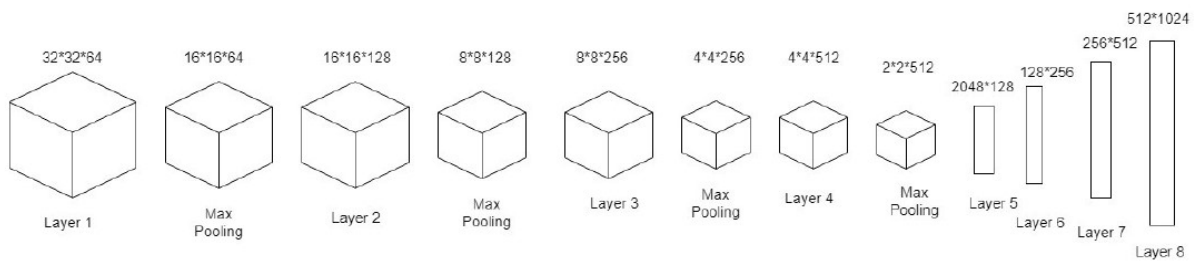

Fig. 4. Our proposed DCT-CompCNN model for CIFAR-10 Dataset.

## III. Experimental Results

As discussed in the previous section, we have done experimentation with both the approaches. In our first approach, we have quantized JPEG Compressed Image, which is partially decoded as well as dequantized, and then after some transformation it is fed to the network. Whereas in our second approach, we have an un-quantized Compressed JPEG image, which is partially decoded, and after some transformation is fed to the network. Both quantized and un-quantized compressed JPEG images for Dogs vs Cats and CIFAR-10 datasets were generated. Now, these two sets of compressed Dogs vs Cats dataset was applied separately on the existing ResNet-50 model for training. But before training the quantized compressed JPEG image, every image of Dog vs Cat dataset were resized to 224 x 224, and now partial decoding along with de-quantization were applied on them as shown in Figure 2, and after performing some transformation i.e. up-sampling or down-sampling on obtained DCT coefficients, is fed to the network. But in case of un-quantized compressed JPEG image, there is no de-quantization operation after partial decoding, which is shown in Figure 3, is fed into the ResNet-50 Network. After training the ResNet-50 model, both approaches give a satisfactory result. As ResNet-50 is a well-known model which also reduces the problem of vanishing gradient. So, a better accuracy and training speed is obvious in both the approaches. So, to verify whether any other random model will give a good accuracy or not, we proposed a model DCT-CompCNN as shown in Figure 4, which was again trained on two sets of compressed CIFRAR-10 dataset, separately, which has more classes than Dogs vs Cat Dataset. We trained the model following the same approaches as we did in ResNet-50 and the satisfactory results were obtained. Throughout the whole experiment, we have used a learning rate of 0.01 which reduces 10 percent after every 10 epochs, the proposed model was trained on 100 epochs, with a batch size of 128, while ResNet-50 model was trained on 45 epochs with a batch size of 32. Adam Optimiser was used during training, to leverage the power of adaptive learning rate methods to find individual learning rate for each parameter. Batch normalisation was used during training, for improving the speed performance and maintain the stability of neural network

TABLE I : CLASSIFICATION RESULT AND INFERENCE SPEED (IN SEC) FOR RESNET-50 MODEL TRAINED ON DOGS VS CATS DATASET

| Type of Operation | Accuracy(%) | Inference Speed |
|---|---|---|
| Quantized (Downsampling) | 81.63 | 880 ± 5 |
| Quantized (Upsampling) | 83.56 | 1072 ± 5 |
| Unquantized (Downsampling) | 82.43 | 874 ± 5 |
| Unquantized (Upsampling) | 84.78 | 1084 ± 5 |

TABLE II : CLASSIFICATION RESULT AND INFERENCE SPEED (IN SEC) FOR PROPOSED DCT-CompCNN MODEL TRAINED ON CIFAR-10 DATASET

| Type of Operation | Accuracy(%) | Inference Speed |
|---|---|---|
| Quantized (Downsampling) | 81.45 | 106 ± 3 |
| Quantized (Upsampling) | 85.95 | 145 ± 3 |
| Unquantized (Downsampling) | 84.13 | 110 ± 3 |
| Unquantized (Upsampling) | 86.31 | 162 ± 3 |

The performances of two proposed approaches have been shown in Table I and Table II. Both the tables depicts the accuracy and the inference speed (in sec for one epoch) for the proposed DCT-CompCNN and ResNet-50 models on different types of operation performed on JPEG Compressed image i.e. quantized JPEG image with up-sampling as transformation, quantized JPEG image with down-sampling, un-quantized JPEG image with up-sampling and un-quantized JPEG image with down-sampling. Table I, depicts the accuracy and inference speed (in sec for one epoch) for different operations on ResNet-50 model, trained on Dogs vs Cat dataset. The highest accuracy obtained in 45 epochs is 84.78% in case of un-quantized image with up-sampling as transformation, while the least inference speed (in sec) of 874±5 was achieved in case of un-quantized image with down-sampling. Table II depicts the accuracy and inference speed (in sec) for one epoch of different operations for our proposed DCT-CompCNN model, the highest accuracy obtained in 100 epochs is 86.31% in case of un-quantized image with up-sampling as transformation, while the least inference speed (in sec) of 106±3 was achieved in case of quantized image with down-sampling. Thus we obtained a good accuracy even on our proposed model also, which verifies that the proposed approach works well on the random model also. Analysing the result we observed that there is less difference in accuracy in case of quantized JPEG image and un-quantized JPEG image, which shows that low frequency regions are more responsible for learning the features of an image as compare to that of high frequency regions. We also analysed that when we down-sampled the luma channel while transformation, we achieved less inference speed, but higher error, which is vice versa in case of up-sampled chroma channels.

## IV. Conclusion

In this paper, we proposed and tested our idea on modifying the input representation for training the existing ResNet-50 and our proposed DCT-CompCNN model for image classification problem. We proposed two approaches for input to the network. We already have compressed JPEG images in our first approach, we apply partial decoding and de-quantization on it, and after applying some transformation on the DCT Coefficients obtained, is fed as input to the neural network. While in the second approach, the compressed JPEG image was not quantized during compression, so we have un-quantized JPEG compressed image, and we apply partial decoding on it, and after some transformation, the obtained matrix is fed as input to the network. Both the approaches were applied on ResNet-50 and proposed DCT-CompCNN model and a good accuracy was observed.